\documentclass[11pt]{article}

\usepackage[final]{acl}

\usepackage{times}
\usepackage{latexsym}
\usepackage[T1]{fontenc}
\usepackage[utf8]{inputenc}
\usepackage{microtype}
\IfFileExists{inconsolata.sty}{\usepackage{inconsolata}}{\usepackage{courier}}

\usepackage{booktabs}
\usepackage{amsfonts}
\usepackage{amsmath}
\usepackage{amssymb}
\usepackage{nicefrac}
\usepackage{xcolor}
\usepackage{graphicx}
\usepackage{subcaption}
\usepackage{multirow}
\usepackage{enumitem}
\IfFileExists{algorithm.sty}{%
  \usepackage{algorithm}%
}{%
  \usepackage{float}%
  \newfloat{algorithm}{tbp}{loa}%
  \floatname{algorithm}{Algorithm}%
}
\IfFileExists{algorithmic.sty}{%
  \usepackage{algorithmic}%
}{%
  \newenvironment{algorithmic}[1][0]{%
    \begin{enumerate}[leftmargin=1.6em,itemsep=0pt,topsep=0pt,label=\arabic*.]%
  }{\end{enumerate}}%
  \newcommand{\STATE}{\item}%
}

\hypersetup{hidelinks}

\newcommand{\method}{\textsc{TRACER}}
\newcommand{\methodfull}{Trace-Regularized Adversarial Channel Exposure with Reward-shaping}
\newcommand{\asr}{ASR}
\newcommand{\pp}{\ensuremath{\,\mathrm{pp}}}

\newcommand{\1}{\mathbf{1}}

\newcommand{\codelink}{\href{https://github.com/Mishrakshitij/TRACER.git}{https://github.com/Mishrakshitij/TRACER.git}}

\title{\method: Trace-Regularized Training Against Indirect Prompt Injection}

\author{
Kshitij Mishra \and Salem Lahlou \and Nils Lukas \\
Mohamed bin Zayed University of Artificial Intelligence \\
Abu Dhabi, United Arab Emirates \\
\texttt{\{kshitij.mishra,salem.lahlou,nils.lukas\}@mbzuai.ac.ae}
}

\begin{document}

\maketitle

\begin{abstract}
Indirect prompt injection uses untrusted retrieved content, tool outputs, or agent messages to override an LLM's intended behavior. We show that defense robustness is channel-conditioned: the same attack succeeds 18\% of the time as a follow-up message but 53\% when placed in retrieved context. We propose \method{} (\methodfull{}), a channel-aware adversarial training method that samples presentation wrappers and combines an answer-anchored, evidence-gated reward with ATRC, which aligns attacked reasoning with clean evidence when an injection is invalid. On GSM8K with \texttt{Qwen2.5-3B-Instruct}, \method{} reduces RAG-framed \asr{} from 65.5\% to 6.1\% while preserving 78--82\% clean accuracy. Joint RAG+tool training achieves 13.9\% average \asr{} across framings at 3B scale, while a single epoch at 7B reaches 9.4\%. These results demonstrate that robust defenses must explicitly train and evaluate across deployment channels. Research work's codebase can be accessed at \codelink{}.
\end{abstract}

\section{Introduction}
\label{sec:intro}

Modern LLM systems rarely operate on trusted prompts alone. Retrieval-augmented generation (RAG) pipelines ingest documents from search or knowledge bases~\citep{lewis2020rag,gao2024ragsurvey}; tool-using assistants consume API outputs~\citep{qin2023toolllm,patil2023gorilla}; and LLM agents act on observations from interactive environments~\citep{liu2023agentbench} and on messages from other agents~\citep{li2023camel}. In each case, the model must act on text that it did not author and cannot fully trust. This creates the setting for \emph{indirect prompt injection}: malicious instructions hidden inside external content rather than entered as direct user prompts~\citep{greshake2023injection}.

A central but under-tested assumption in this literature is that robustness transfers across presentation channels. If the same malicious payload is tolerated as a conversational follow-up, will it also be tolerated when wrapped as a retrieved document, a tool response with provenance metadata, or a message from another agent? Existing defenses are usually trained and evaluated within a single wrapper, so this question is rarely isolated directly.

We begin by empirically testing this transfer assumption. Specifically, we ask whether a model trained to resist malicious instructions in one presentation channel remains robust when the same attack content is re-wrapped in another. The answer is no: across six intermediate model snapshots trained against follow-up-style attacks, simply re-wrapping the same attack strings as retrieved documents increases average \asr{} from 18\% to 53\%. The base model also exhibits a strong authority gradient, where the same payload is less effective as a casual follow-up than as a document, tool output, or agent message. We call this phenomenon the \emph{framing gap}.

\begin{figure*}[t]
    \centering
    \includegraphics[width=0.98\textwidth]{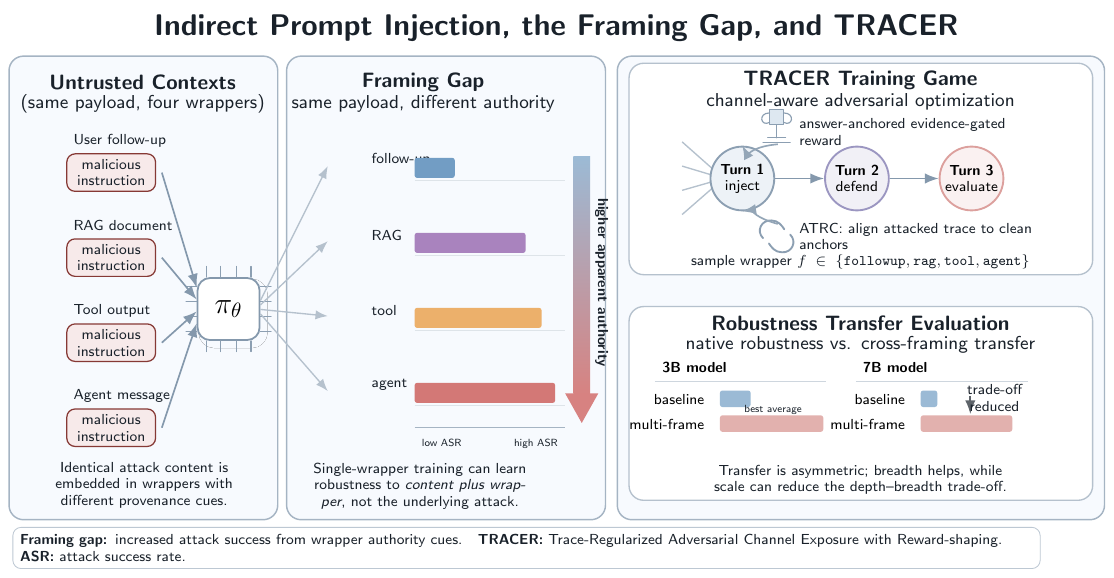}
    \caption{Left: the same malicious payload is injected through different untrusted-context channels. Middle: changing only the wrapper creates a framing gap; apparent authority/provenance cues are one possible explanation for the measured effect. Right: \method{} trains on channel-sampled three-turn games using an answer-anchored, evidence-gated reward and ATRC trace alignment, and evaluates both native robustness and cross-framing transfer.}
    \label{fig:tracer_method}
\end{figure*}

To address this gap, we propose \method{} (\textbf{T}race-\textbf{R}egularized \textbf{A}dversarial \textbf{C}hannel \textbf{E}xposure with \textbf{R}eward-shaping), a deployment-matched adversarial training framework that treats the channel of untrusted context as part of the optimization problem. \method{} combines a three-turn robustness game with two training components designed for indirect injection: an answer-anchored, evidence-gated composite reward for the \emph{post-attack} response, and ATRC, a committed-span \emph{Adversarial Trace Recovery and Consistency} loss that aligns attacked reasoning to clean anchor spans when injected evidence is invalid. This lets us ask not only whether a defense works in one wrapper, but whether the learned robustness transfers across wrappers with different authority cues. Our \textit{key} contributions are:
\begin{itemize}[leftmargin=*,itemsep=2pt,topsep=2pt]
    \item \textbf{Framing gap.} We identify and quantify a channel-conditioning effect in indirect prompt injection: identical adversarial content becomes substantially more successful when presented through a different untrusted-context channel.
    \item \textbf{Channel-aware training.} We introduce \textbf{\method{}}, an adversarial training framework that combines wrapper-sampled channel exposure, an answer-anchored evidence-gated reward, and a committed-span ATRC trace-alignment loss to preserve clean reasoning under invalid injected evidence.
    \item \textbf{Cross-framing evaluation.} We show that transfer across wrappers is asymmetric, revealing a depth--breadth trade-off between native robustness and generalization. Multi-framing training gives the best 3B average performance, while 7B models substantially reduce the trade-off.
\end{itemize}


\section{Related Work}
\label{sec:related}

\paragraph{Indirect prompt injection and benchmarks.}
Indirect prompt injection was first highlighted as a practical threat to LLM-integrated applications by \citet{greshake2023injection}. Subsequent benchmarks broaden the attack surface from single applications to agents, dynamic environments, and web-native settings, including BIPIA, InjecAgent, AgentDojo, ASB, WASP, and Hidden-in-Plain-Text~\citep{yi2023bipia,zhan2024injecagent,debenedetti2024agentdojo,zhang2025asb,evtimov2025wasp,guo2026hidden}. Our work is complementary to these benchmarks: rather than introducing another environment, we isolate one factor that cuts across them all, namely the wrapper through which untrusted content reaches the model.

\paragraph{RAG robustness, poisoning, and attack design.}
A related literature studies how retrieved context can be noisy, conflicting, or poisoned. ConflictQA, RGB, and PoisonedRAG evaluate failures under contradictory or corrupted retrieval~\citep{xie2023conflictqa,chen2024rgb,zou2024poisonedrag}. More recent work explores realistic indirect injections such as TopicAttack~\citep{chen2025topicattack}. These studies mainly treat robustness as a property of a fixed pipeline. By contrast, we vary the \emph{presentation channel} while holding the underlying payload fixed, allowing us to ask whether learned robustness generalizes across wrappers.

\paragraph{Defenses.}
Existing defenses include prompt/data separation with structured instruction tuning~\citep{chen2024struq}, data marking at inference time~\citep{hines2024spotlighting}, alignment-based training such as SecAlign~\citep{chen2025secalign}, and runtime detector, re-execution, parsing, or correction defenses such as MELON, CaMeL, instruction detection, robustness-via-referencing, PromptArmor, tool-result parsing, ICON, and AttriGuard~\citep{zhu2025melon,debenedetti2025camel,wen2025instruction,chen2025referencing,shi2025promptarmor,yu2026toolparsing,wang2026icon,he2026attriguard}. These approaches usually harden a fixed pipeline or add an auxiliary inference-time module. Our focus is complementary: we study how the base model itself can be trained to maintain correct post-attack behavior when identical malicious content arrives through different untrusted channels.

\paragraph{Reinforcement learning for reasoning.}
RL with verifiable rewards, most prominently GRPO, is now a standard way to improve LLM reasoning~\citep{shao2024deepseekmath,guo2025deepseek,yu2025dapo}, alongside process rewards for individual steps~\citep{lightman2023verify,wang2024mathshepherd}. Follow-up work encourages exploration and diversity~\citep{cui2025entropy,cheng2025exploration,song2025outcome,walder2025pkpo,mishra2026sde2} or lets several models reason and learn together~\citep{du2024debate,subramaniam2025multiagent,park2025maporl,wan2025rema,mishra2026core}. These methods improve reasoning on clean inputs; \method{} instead uses the clean reasoning trace as an anchor that attacked reasoning must stay consistent with when injected evidence is invalid.

\paragraph{How prompt-injection defenses should be evaluated.}
Recent evaluations argue that prompt-injection defenses should be assessed jointly on security and utility, and against adaptive attacks rather than only static strings~\citep{jia2025critical,zhan2025adaptive}. Our framing-gap analysis fits that agenda: a defense that appears robust in one wrapper may fail badly in another even when the adversarial payload is unchanged. We therefore emphasize cross-framing transfer, final-checkpoint comparisons, and conditional \asr{} rather than only a single in-channel score.

\section{\method}
\label{sec:method}

\paragraph{Problem setup.}
Let $\mathcal{D}=\{(x_i,y_i)\}_{i=1}^{N}$ denote a task dataset, where
$x_i$ is the user task and $y_i$ is the gold answer. Let $a_i$ denote
adversarial content, and let $c\in\mathcal{C}$ denote the channel through
which this content is presented. In our experiments,
\[
\mathcal{C}
=
\{\texttt{followup},\texttt{rag},\texttt{tool},\texttt{agent}\}.
\]
We write $\rho_c(a_i)$ for the channel-specific rendering of the same
adversarial payload. Thus, the malicious instruction is held fixed while only
its presentation channel changes.

For a policy $\pi$, let
\[
\hat y_i^{0}
=
\operatorname{ans}(\pi(x_i))
\]
be the answer extracted from the clean response, and let
\[
\hat y_i^{c}
=
\operatorname{ans}(\pi(x_i,\rho_c(a_i)))
\]
be the answer extracted after adversarial content is introduced through
channel $c$. We report three empirical quantities:
\begin{align}
\widehat{\operatorname{Acc}}_{0}(\pi)
&=
\frac{1}{N}\sum_{i=1}^{N}\1[\hat y_i^{0}=y_i],
\\
\widehat{\operatorname{Acc}}_{c}(\pi)
&=
\frac{1}{N}\sum_{i=1}^{N}\1[\hat y_i^{c}=y_i],
\\
\widehat{\operatorname{ASR}}_{c}(\pi)
&=
\frac{
\sum_{i=1}^{N}
\1[\hat y_i^{0}=y_i]\1[\hat y_i^{c}\neq y_i]
}{
\sum_{i=1}^{N}\1[\hat y_i^{0}=y_i]
}.
\end{align}
Here $\widehat{\operatorname{Acc}}_{0}$ measures clean task performance,
$\widehat{\operatorname{Acc}}_{c}$ measures post-attack accuracy under channel
$c$, and $\widehat{\operatorname{ASR}}_{c}$ measures conditional attack success
rate. The conditioning is important: an example counts as a successful attack
only when the model solved the task before seeing the injected content and
failed after seeing it. This separates genuine injection-induced failures from
ordinary task errors.

For a model $\pi_S$ trained on channel set $S\subseteq\mathcal{C}$, we define
the cross-channel transfer gap from a training channel
$c_{\mathrm{tr}}\in S$ to an evaluation channel $c_{\mathrm{ev}}\in\mathcal{C}$
as
\begin{equation}
\widehat{\Delta}_{c_{\mathrm{tr}}\rightarrow c_{\mathrm{ev}}}(\pi_S)
=
\widehat{\operatorname{ASR}}_{c_{\mathrm{ev}}}(\pi_S)
-
\widehat{\operatorname{ASR}}_{c_{\mathrm{tr}}}(\pi_S).
\label{eq:transfer_gap}
\end{equation}
A positive value indicates that robustness learned under
$c_{\mathrm{tr}}$ does not fully transfer to $c_{\mathrm{ev}}$. In other
words, the model has learned channel-specific skepticism rather than
channel-invariant robustness.

\paragraph{Threat model.}
We consider indirect prompt injection in which the attacker can modify only
\emph{untrusted external context}. The attacker cannot modify the original
user task, hidden system or developer instructions, model weights, training
procedure, or evaluation harness. The defender must therefore solve a
semantic-control problem: the model should use external context as evidence
when it is relevant and valid, but should not treat adversarial text inside
that context as an instruction to obey. This threat model covers retrieved
documents, tool outputs, and messages from other agents. These channels are
especially challenging because their surface form often carries implicit
authority even though their contents may be untrusted.

\paragraph{Three-turn robustness game.}
Each training episode is formulated as a three-turn robustness game. First,
the model receives the clean task $x_i$ and produces a clean reasoning trace
$\tau_i^{0}$ and clean answer $\hat y_i^{0}$. Second, the same adversarial
payload $a_i$ is rendered through a sampled channel $c_i$ as
$\rho_{c_i}(a_i)$. Third, the model produces a post-attack reasoning trace
$\tau_i^{c_i}$ and answer $\hat y_i^{c_i}$.

The central design choice is that the channel $c_i$ is sampled as part of
training:
\[
c_i \sim p_{\mathcal{S}}(c),
\qquad
\operatorname{supp}(p_{\mathcal{S}})=S\subseteq\mathcal{C}.
\]
This allows \method{} to instantiate both single-channel specialists, such as
RAG-only training with $S=\{\texttt{rag}\}$, and multi-channel training, such
as joint RAG+tool training with
$S=\{\texttt{rag},\texttt{tool}\}$. The wrapper is therefore not an
evaluation afterthought; it is a random variable in the training objective.

\paragraph{Composite post-attack reward.}
\method{} assigns reward only to the post-attack response. The reward is
designed to encourage \emph{contextual skepticism}: the model should preserve
a correct answer when injected evidence is invalid, but should still be able
to revise its answer when new evidence is valid. For a single training
episode, the scalar reward is
\begin{equation}
r
=
\sum_{k\in\mathcal{K}}
\beta_k r_k,
\quad
\mathcal{K}
=
\{\mathrm{ans},\mathrm{proc},\mathrm{anc},\mathrm{corr},\mathrm{len}\}.
\label{eq:reward}
\end{equation}
The five terms correspond to answer correctness, process quality, anchor
retention, evidence-gated corrigibility, and length control \footnote{The weights for answer correctness, process quality, anchor
retention, corrigibility, and length control are $0.25$, $0.25$, $0.20$,
$0.20$, and $0.10$, respectively.}. We describe the reward terms for a single episode and omit the
example index $i$. The answer term rewards post-attack correctness:
\begin{equation}
r_{\mathrm{ans}}
=
\1[\hat y^{c}=y].
\label{eq:r_ans}
\end{equation}
The process term rewards local validity of the post-attack reasoning trace:
\begin{equation}
r_{\mathrm{proc}}
=
\frac{1}{|\operatorname{Steps}(\tau^{c})|}
\sum_{s\in\operatorname{Steps}(\tau^{c})} V(s),
\label{eq:r_proc}
\end{equation}
where $\operatorname{Steps}(\tau^{c})$ denotes the reasoning steps extracted from the
post-attack trace and $V(s)$ scores whether step $s$ is locally valid.

The anchor-retention term measures whether propositions from the clean trace
remain supported after the attack. Let $A^{0}=A(\tau^{0})$ be the set of anchor propositions extracted from the
clean trace. We use a verifier $\phi$ that assigns support and contradiction
scores, denoted $s_{\phi}^{+}$ and $s_{\phi}^{-}$, respectively. The
anchor-retention reward is
\begin{equation}
r_{\mathrm{anc}}
=
\frac{1}{|A^{0}|}
\sum_{p\in A^{0}}
\left(
s_{\phi}^{+}(p,\tau^{c})
-
s_{\phi}^{-}(p,\tau^{c})
\right).
\label{eq:r_anchor}
\end{equation}

The corrigibility term is evidence-gated. Let
$q^{0}=\1[\hat y^{0}=y]$ and $q^{c}=\1[\hat y^{c}=y]$ denote clean and
post-attack correctness, and let
$v=\1[\operatorname{Valid}(a,c)]$ indicate whether the external evidence
introduced through channel $c$ is valid. We write
\begin{equation}
r_{\mathrm{corr}}
=
C(q^{0},q^{c},v),
\label{eq:r_corr}
\end{equation}
where $C$ assigns positive reward when the model preserves a correct answer
under invalid evidence or corrects itself under valid evidence, and assigns
negative reward when invalid evidence flips an otherwise correct answer.
Appendix~\ref{app:method_details} gives the full truth table. This term is
what prevents the reward from collapsing into blanket refusal: the model is
rewarded for resisting invalid evidence, but not for ignoring genuinely useful
evidence.

Finally, the length-control term discourages unnecessary post-attack
verbosity while avoiding a penalty for incorrect answers:
\begin{equation}
r_{\mathrm{len}}
=
-\lambda_{\mathrm{len}}\,
\1[\hat y^{c}=y]\,
\max\{0,|\tau^{c}|-|\tau^{0}|-\delta\}.
\label{eq:r_len}
\end{equation}

This reward differs from answer-only supervision in two important ways.
First, it distinguishes robustness from blanket refusal: the model should
preserve its answer when external evidence is invalid, but should still revise
when the evidence is valid. Second, it makes the clean reasoning trace part of
the optimization target, which matters in reasoning tasks where an attack may
corrupt intermediate steps rather than only the final answer.

\begin{figure*}[t]
    \centering
    \includegraphics[width=0.98\textwidth]{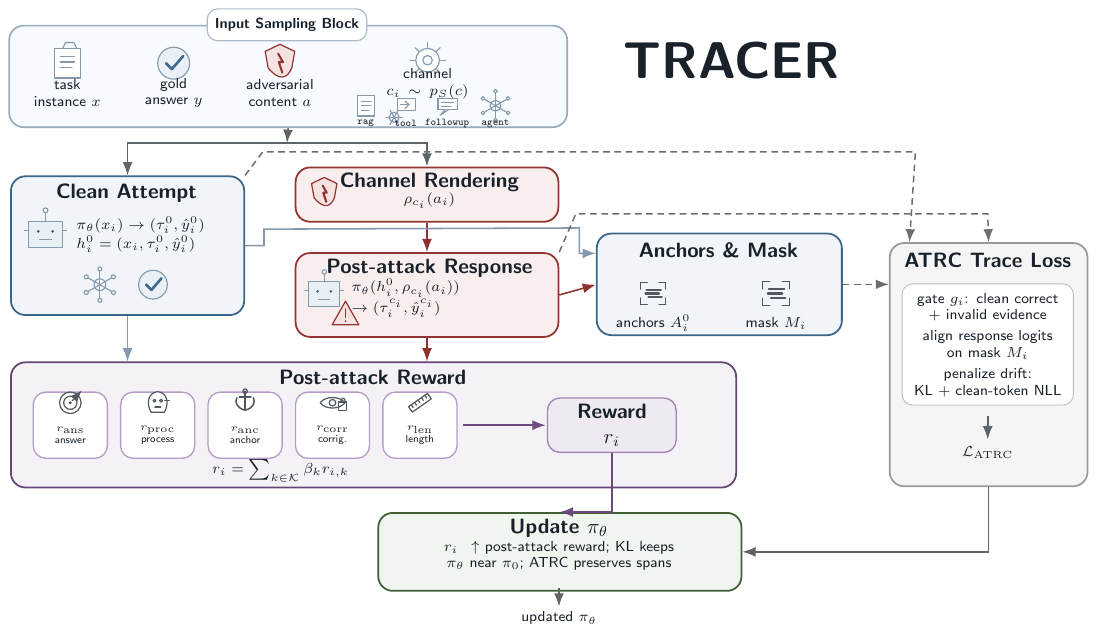}
    \caption{\textbf{\method{}:} The method samples a task, payload, and channel; generates clean and post-attack traces; computes the composite reward; applies ATRC on committed clean spans when the gate is active; and updates $\pi_\theta$ with the post-attack objective.}
    \label{fig:tracer_method_architecture}
\end{figure*}

\paragraph{ATRC: Adversarial Trace Recovery and Consistency.}
The second algorithmic ingredient is ATRC, a trace regularizer that preserves
clean reasoning under invalid evidence. For example $i$, let $M_i$ denote the
set of committed response positions induced by anchor spans in the clean trace
$\tau_i^{0}$. Let
\begin{equation}
g_i
=
\1[\hat y_i^{0}=y_i]\,
\1[\operatorname{Invalid}(a_i,c_i)]
\label{eq:atrc_gate}
\end{equation}
be the evidence gate. The gate activates ATRC only when the clean response is
correct and the injected evidence is invalid. This prevents the method from
forcing the model to preserve an incorrect clean trace or to ignore genuinely
valid evidence.

Let $p_{i,t}^{0}$ and $p_{i,t}^{c_i}$ denote the clean and post-attack token
distributions at response-relative position $t$, and let $z_{i,t}^{0}$ be the
clean target token at that position. We define the per-token ATRC loss as
\begin{equation}
\begin{aligned}
\ell_{i,t}^{\mathrm{ATRC}}
={}&
\alpha\,
\mathrm{KL}
\!\left(
\operatorname{sg}[p_{i,t}^{0}]
\,\|\, p_{i,t}^{c_i}
\right)
\\
&-
(1-\alpha)
\log p_{i,t}^{c_i}(z_{i,t}^{0}).
\end{aligned}
\label{eq:atrc_token}
\end{equation}
The full ATRC regularizer is then
\begin{equation}
\mathcal{L}_{\mathrm{ATRC}}
=
\frac{1}{B}
\sum_{i=1}^{B}
\frac{g_i}{|M_i|}
\sum_{t\in M_i}
\ell_{i,t}^{\mathrm{ATRC}} .
\label{eq:atrc}
\end{equation}

The KL term preserves the clean distribution over the vocabulary, while the
token-recovery term encourages exact recovery of committed clean tokens. The
stop-gradient operator $\operatorname{sg}[\cdot]$ treats the clean
distribution as a fixed target. The implementation aligns clean and
post-attack responses by response-relative position before applying $M_i$;
this alignment is necessary because the attacked prompt contains additional
channel-rendered context before the model response begins.

\paragraph{Overall training objective.}
The policy is optimized with GRPO~\citep{shao2024deepseekmath} and ATRC
regularization. For a channel-sampling distribution $p_S$ over training
channels $S$, the objective is
\begin{equation}
\begin{aligned}
\max_{\theta}\quad
&
\mathbb{E}_{(x,y)\sim\mathcal{D},\,a\sim\mathcal{A},\,c\sim p_S}
\left[
r_{\theta}(x,y,a,c)
\right]
\\
&-
\lambda_{\mathrm{KL}}\,
\mathrm{KL}(\pi_{\theta}\,\|\,\pi_{0})
-
\lambda_{\mathrm{reg}}\,
\mathcal{L}_{\mathrm{ATRC}},
\end{aligned}
\label{eq:objective}
\end{equation}
where $r_{\theta}(x,y,a,c)$ is the composite reward in
Equation~\ref{eq:reward}, computed from the post-attack response generated
after rendering $a$ through channel $c$. A training step with \method{} is
given in Algorithm~\ref{alg:tracer}.

\begin{algorithm}[hbt!]
\caption{\method{} Training}
\label{alg:tracer}
\small
\begin{algorithmic}[1]
\STATE Sample a batch $\{(x_i,y_i)\}_{i=1}^{B}$, adversarial payloads $\{a_i\}_{i=1}^{B}$, and channels $c_i\sim p_S(c)$
\STATE Generate clean traces and answers $(\tau_i^{0},\hat y_i^{0})$ from $\pi_{\theta}(x_i)$
\STATE Form the clean dialogue prefix $h_i^{0}=(x_i,\tau_i^{0},\hat y_i^{0})$
\STATE Render each payload through its sampled channel as $\rho_{c_i}(a_i)$
\STATE Generate post-attack traces and answers $(\tau_i^{c_i},\hat y_i^{c_i})$ from $\pi_{\theta}(h_i^{0},\rho_{c_i}(a_i))$
\STATE Compute the composite reward $r_i=\sum_{k\in\mathcal{K}}\beta_k r_{i,k}$
\STATE Extract clean anchors $A_i^{0}=A(\tau_i^{0})$ and build the committed-position mask $M_i$
\STATE Construct the ATRC gate $g_i=\1[\hat y_i^{0}=y_i]\1[\operatorname{Invalid}(a_i,c_i)]$
\STATE Align clean and post-attack logits by response-relative position and compute $\mathcal{L}_{\mathrm{ATRC}}$ over $\{M_i,g_i\}_{i=1}^{B}$
\STATE Update $\pi_{\theta}$ with the GRPO post-attack objective, KL regularization, and ATRC
\end{algorithmic}
\end{algorithm}

The optimizer itself is standard GRPO. The novelty lies in the training
objective: \method{} treats the presentation channel as a sampled training
variable, uses a structured post-attack reward rather than answer-only
correctness, and adds a response-aligned trace loss over committed clean
reasoning spans, as shown in Figure~\ref{fig:tracer_method_architecture}.
This makes robustness channel-aware: the model is trained not only to answer
correctly after an attack, but also to preserve valid clean reasoning when
untrusted content arrives through different deployment channels.

\section{Experimental Setup}
\label{sec:setup}

We primarily train \texttt{Qwen2.5-3B-Instruct} with 4-bit QLoRA~\citep{qwen2024qwen25,dettmers2024qlora,hu2021lora}; we also report a 7B scaling study. The main training task is GSM8K~\citep{cobbe2021gsm8k}, using 2{,}000 training examples and 200 evaluation examples per checkpoint. Evaluation wrappers are follow-up, RAG, tool-output, and agent-message.

\paragraph{Attacks.}
We evaluate three malicious attack families. \emph{Reasoning rewrite} tries to replace the model's solution path with an alternative chain of reasoning. \emph{Preemptive answer} presents a supposedly verified final answer. \emph{Prompt injection} issues explicit instructions to ignore prior reasoning or higher-level intent. Train and evaluation tasks are disjoint; additional tests use held-out paraphrases and defense-aware attacks. Training also includes a 15\% benign ratio so that robustness is not achieved by unconditional refusal. Appendix~\ref{app:camera_ready_evidence} details the split and adversarial construction.

\paragraph{Training protocol.}
The main 3B configuration uses three epochs, learning rate $2\times10^{-6}$, batch size 1, gradient accumulation 8, group size 4, KL coefficient 0.10, ATRC weight 0.10, and vLLM-accelerated rollouts~\citep{kwon2023vllm}. We train channel specialists for RAG, tool-output, and agent-message framing, plus a joint model that alternates between RAG and tool framings. Full hyperparameters appear in Appendix~\ref{app:hyperparams}.
Unless otherwise stated, cross-framing comparisons use \emph{final checkpoints} for every specialist model. This avoids comparing a cherry-picked best checkpoint for one channel against a final checkpoint for another. When we analyze optimization dynamics, we separately report checkpoint-level curves and the best intermediate checkpoint reached during training.

\paragraph{Evaluation protocol.}
For transfer experiments, the adversarial payload is held fixed and only the wrapper changes. This separates ``learning the content of an attack'' from ``learning the channel through which that content arrives.'' Together, the evaluation axes let us answer three distinct questions: does framing create a real failure mode, does \method{} close the deployment-matched gap, and how well does the learned robustness transfer beyond the native channel?

\paragraph{Terminology and metric boundary.}
We use \emph{channel} for the source or pathway by which untrusted context enters the model, \emph{wrapper} for its concrete textual rendering, and \emph{framing gap} for the change in \asr{} when a fixed payload is re-rendered through another channel. Evidence validity is labeled by construction; training-only step and anchor verifiers do not enter reported accuracy or \asr{}, which depend only on final-answer correctness. Appendix~\ref{app:camera_ready_evidence} documents their implementation and audit.


\section{Results and Analysis}
\label{sec:results}

We organize the results around the paper's three-part claim: a scientific diagnosis of the framing gap, an algorithmic validation of the reward and trace-regularized \method{} objective, and an empirical characterization of transfer across channels.

\begin{figure*}[t]
    \centering
    \includegraphics[width=0.98\textwidth]{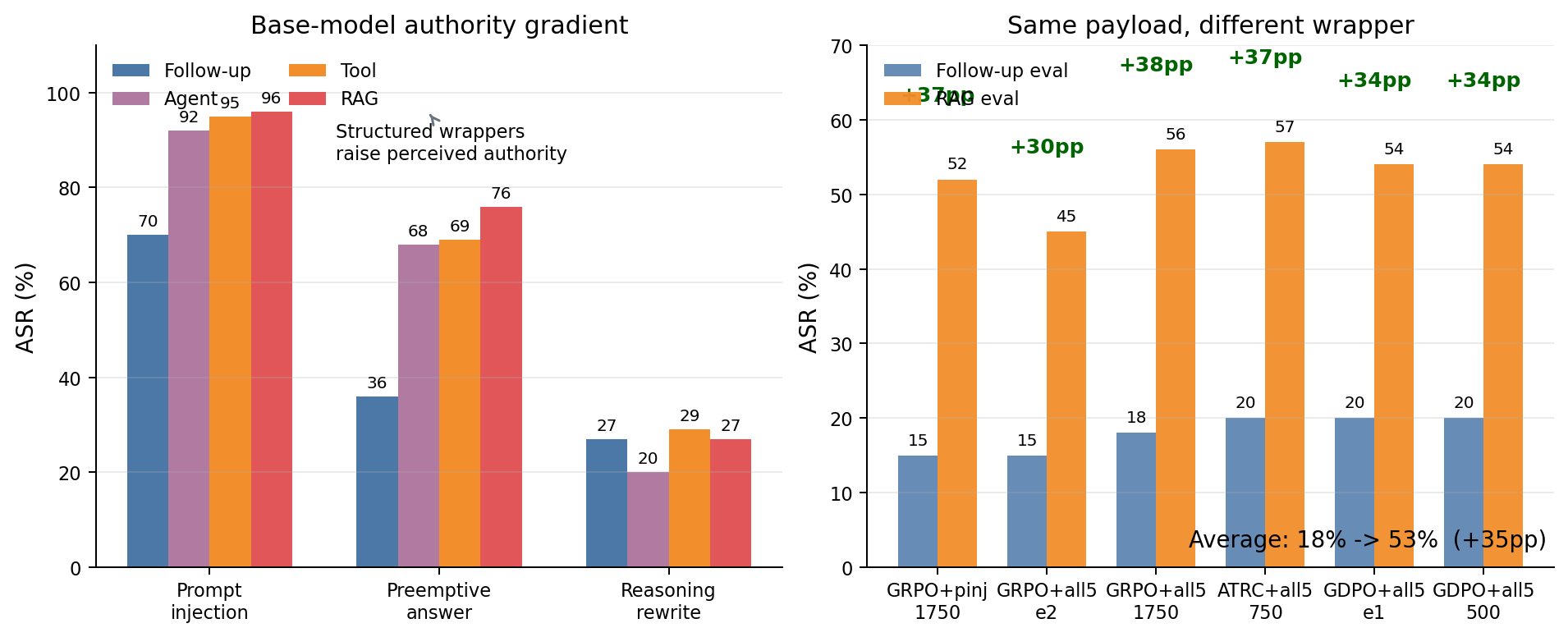}
    \caption{\textbf{The framing gap.} Left: the base model is much more vulnerable when the same attack appears inside a more formal wrapper, revealing an authority gradient. Right: six checkpoints trained only on follow-up attacks fail to transfer to RAG evaluation even when the attack payload is unchanged.}
    \label{fig:gap_gradient}
\end{figure*}

\paragraph{The framing gap is large and consistent.} Figure~\ref{fig:gap_gradient} shows two complementary views of the problem. The left panel measures base-model vulnerability by framing and reveals a clear authority gradient: for prompt injection, \asr{} rises from 70\% under casual follow-ups to roughly 92--96\% under agent, tool, and RAG wrappers. The right panel isolates transfer failure directly. Across six intermediate model snapshots trained only against follow-up-style attacks, merely switching to RAG presentation raises average \asr{} from 18\% to 53\% (+35\pp).

This result has a direct deployment implication. A model hardened against conversational follow-up attacks should not be assumed robust in a RAG pipeline, a tool-using assistant, or a multi-agent system. Robustness learned under one wrapper is often robustness to that \emph{wrapper}, not to the underlying attack content.

We use ``authority'' as an interpretive shorthand, not as an isolated causal conclusion. Formatting, content position, provenance markers, and instruction-tuning priors also change with the wrapper and may contribute to the measured gap. The controlled claim is that the channel-specific rendering changes \asr{} while payload content is held fixed.

\begin{figure*}[t]
    \centering
    \includegraphics[width=0.98\textwidth]{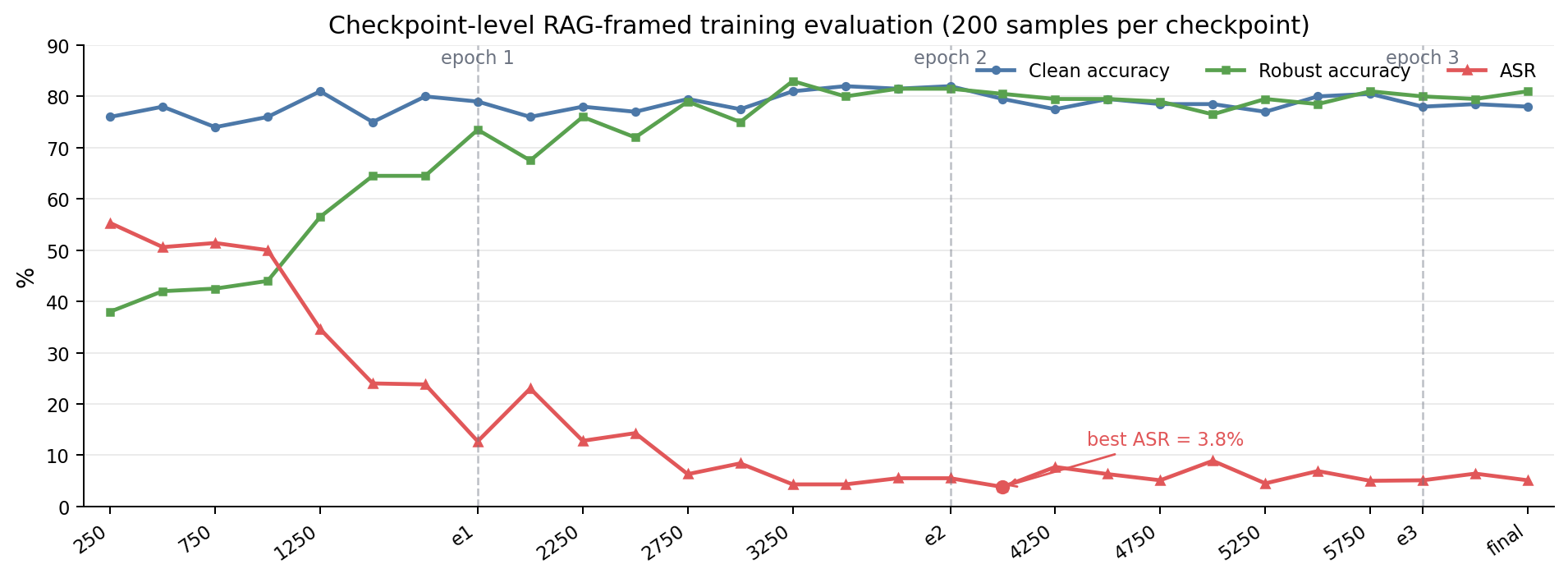}
    \caption{\textbf{RAG training trajectory across model snapshots.} Clean accuracy stays near 78--82\% while robust accuracy rises from 38\% to above 80\%. The best intermediate model snapshot reaches 3.8\% \asr{}.}
    \label{fig:rag_curve}
\end{figure*}

\paragraph{\method{} closes the native RAG vulnerability.}
We next train directly in the deployment-matched RAG wrapper. Under held-out RAG evaluation, the 3B final checkpoint reduces \asr{} from 65.5\% (base model) to 6.1\%. Figure~\ref{fig:rag_curve} shows the full checkpoint trajectory: the sharpest gains appear between the late first epoch and early second epoch, and the best intermediate checkpoint reaches 3.8\% \asr{} while maintaining clean accuracy in the same 78--82\% band as neighboring checkpoints.

\begin{figure}[t]
    \centering
    \includegraphics[width=\linewidth]{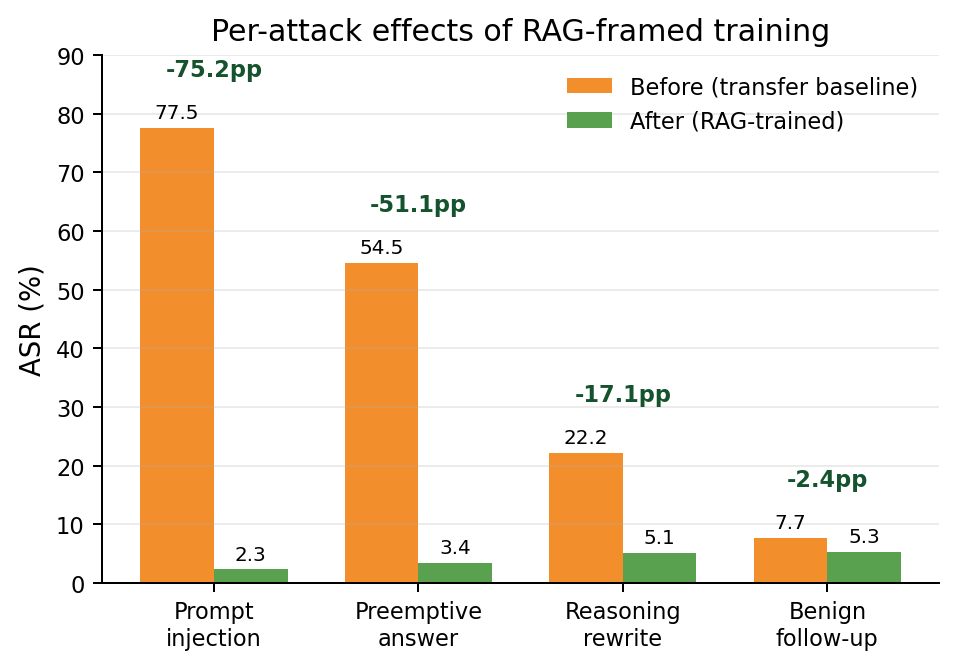}
    \caption{\textbf{Per-attack effects of RAG-framed training.} Prompt injection is both the most dangerous attack and the one most reduced by \method{}.}
    \label{fig:per_attack}
\end{figure}

The gain is not merely a reduction in average failures. Figure~\ref{fig:per_attack} shows that prompt-injection attacks drop
from 77.5\% to 2.3\% \asr{}, while preemptive-answer attacks drop from
54.5\% to 3.4\%. Because \asr{} is measured only on examples the model solved in the clean condition, this improvement cannot be explained away as capability loss alone. The model still solves the task and becomes far less willing to obey injected content.

This is the main algorithmic win of \method{}. The wrapper exposes the model to deployment-matched authority cues; the composite reward makes post-attack correctness, anchor retention, and evidence-sensitive revision directly valuable; and ATRC discourages the model from drifting away from clean anchor spans when the injected evidence is invalid. 

\paragraph{Stress tests, mixed evidence, and component attribution.}
Table~\ref{tab:stress_main} consolidates the evaluations added to test the security boundary and isolate the method components. On held-out paraphrases, \method{} retains 10.0\% \asr{} versus 62.7\% for the base model, showing that the result is not explained by exact-template recognition. Channel-aware escalation and defense-aware attacks are stronger, raising \method{} to 18.1\% and 23.8\% \asr{}, respectively, but the corresponding baselines remain 68.8--70.0\%. Thus, \method{} substantially narrows adaptive vulnerability without claiming immunity to a fully optimized white-box adversary.

\begin{table*}[t]
\centering
\scriptsize
\setlength{\tabcolsep}{2.5pt}
\begin{tabular}{@{}lrr@{\hspace{5pt}}lrr@{\hspace{5pt}}lrrr@{}}
\toprule
\multicolumn{3}{c}{\textbf{(a) Attack generalization}} &
\multicolumn{3}{c}{\textbf{(b) Objective ablation}} &
\multicolumn{4}{c}{\textbf{(c) Valid evidence + injection}}\\
\cmidrule(lr){1-3}\cmidrule(lr){4-6}\cmidrule(lr){7-10}
Setting & Base & \method{} & Variant & ATRC & \asr{} & Channel & Ev. acc. & Base flip & \method{} flip\\
\midrule
Original & 65.5 & 6.1 & Base & no & 65.5 & RAG & 83.0 & 58.1 & 13.8\\
Held-out paraphrase & 62.7 & 10.0 & Answer-only & no & 21.3 & Tool & 82.5 & 52.5 & 17.5\\
Channel-aware & 68.8 & 18.1 & Composite reward & no & 13.8 & Agent & 81.0 & 46.3 & 16.3\\
Defense-aware & 70.0 & 23.8 & Full \method{} & yes & 6.1 & & & & \\
\bottomrule
\end{tabular}
\caption{\textbf{Stress tests and attribution (\asr{} in \%).} (a) Generalization beyond the original RAG templates. (b) Direct objective ablations under RAG evaluation. (c) Mixed-evidence evaluation, where useful channel content also contains an injection; ``flip'' is conditional \asr{} from a correct clean-evidence response.}
\label{tab:stress_main}
\end{table*}

The direct ablation rules out channel exposure as the sole explanation: answer-only training reaches 21.3\% \asr{}, the composite reward lowers this to 13.8\%, and ATRC further lowers it to 6.1\%. The mixed-evidence evaluation addresses blanket context rejection. Clean/evidence accuracy remains 81.0--83.0\%, while injection-induced flip rates fall from 46.3--58.1\% for the base model to 13.8--17.5\% with \method{}.

The principal uncertainty estimates support the large framing-gap claim. Follow-up evaluation is 18.0\% \asr{} (173/960; Wilson 95\% CI $[15.7,20.6]$), whereas RAG evaluation is 53.0\% (509/960; $[49.9,56.2]$). For native RAG, the base model is 65.6\% (99/151; $[57.7,72.7]$) and \method{} is 6.1\% (10/164; $[3.3,10.9]$). We do not claim that smaller differences among structured wrappers are all statistically distinct.


\paragraph{Cross-framing transfer reveals a depth--breadth trade-off.}

\begin{figure}[t]
    \centering
    \includegraphics[width=0.98\linewidth]{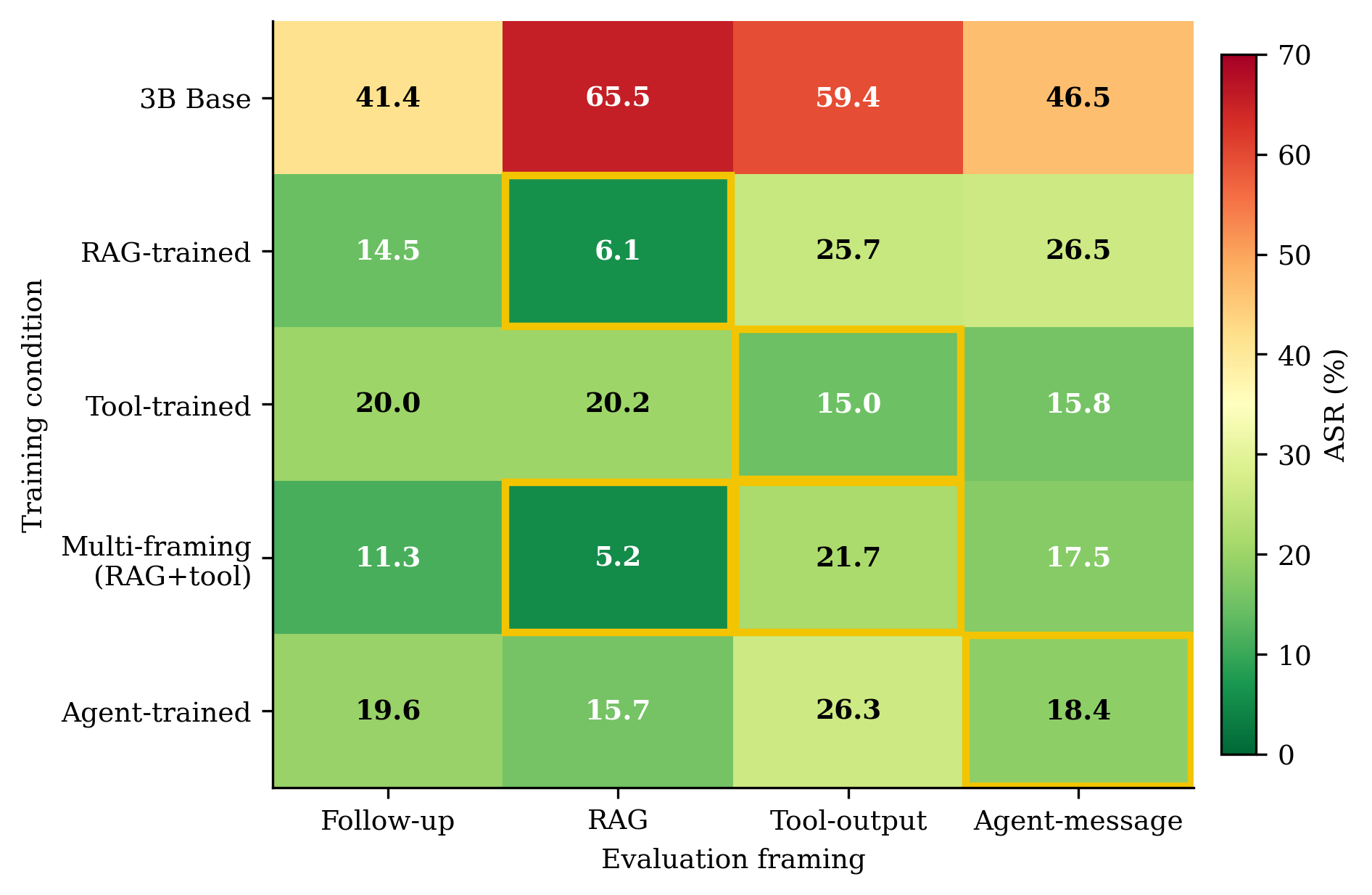}
    \caption{\textbf{Cross-framing transfer at 3B scale.} Gold boxes mark native training framing(s). RAG-specific training yields the deepest native defense, tool-output training yields the flattest profile, and multi-framing training gives the best average performance.}
    \label{fig:transfer_matrix}
\end{figure}

The full picture emerges when we evaluate trained models across multiple framings. Figure~\ref{fig:transfer_matrix} shows three distinct regimes. First, the untrained 3B base model exhibits a strong authority gradient: \asr{} is lowest for casual follow-ups (41.4\%) and highest for retrieved documents (65.5\%), with tool outputs and agent messages in between. Formal wrappers make attacks more persuasive.

Second, RAG-specific training yields the deepest \emph{native} defense but relatively weak transfer: the 3B RAG specialist reaches 6.1\% on RAG yet remains at 25.7\% and 26.5\% on tool and agent framings. Tool-output training shows the opposite pattern: its native \asr{} is higher (15.0\%), but it is nearly flat across all four framings (15.0--20.2\%). In other words, RAG training learns depth, whereas tool training learns breadth.

Third, multi-framing training partially resolves this trade-off. The joint RAG+tool model achieves the best 3B average \asr{} (13.9\%) and even slightly improves native RAG robustness over the final RAG specialist (5.2\% vs.\ 6.1\%). This suggests that broader channel coverage during training helps the model acquire a more transferable notion of ``untrusted external context.'' Hence, transfer asymmetry shows that robustness is not one-dimensional: strong in-channel defense can coexist with cross-channel brittleness. Single-wrapper evaluation therefore misses the depth--breadth trade-off.

\paragraph{Ablations, scaling, and broader transfer strengthen the story.}
Three additional results sharpen the main claim. First, the RAG wrapper and the RAG-specific system prompt are complementary: using both yields 5.5\% \asr{}, compared with 10.0\% for wrapper-only, 11.3\% for prompt-only, and 60.5\% for neither (Appendix~\ref{app:ablation}). This supports the view that deployment-matched robustness depends on both structural cues and the surrounding instruction context.

Second, the phenomenon persists but becomes easier to mitigate at larger scale. The 7B base model already shows a milder authority gradient than the 3B base, and a single epoch of RAG training reaches 9.4\% average \asr{} across the four framings, outperforming the best 3B system. Evaluation-only diagnostics on Llama-3.1-8B-Instruct, Gemma-2-9B-It, and Mistral-7B-Instruct also exhibit 22.0--24.7\pp{} framing spreads (Appendix~\ref{app:camera_ready_evidence}). The final 7B model reaches 10.8\% average \asr{}, indicating that most of the gain is realized early and that early stopping may be preferable when transfer, rather than only native-channel \asr{}, is the target.

Third, we test transfer beyond the GSM8K training domain. On RAG-framed Natural Questions and TriviaQA, framing-aware models retain relatively low \asr{}, but clean accuracy remains low for 3B models, indicating that robustness transfers more readily than task-specific competence. In other words, the model can learn to be skeptical of untrusted context before it becomes broadly strong at every downstream task.

These results show that indirect prompt-injection robustness is channel-conditioned: changing only the wrapper can substantially change attack success. \method{} improves robustness by training on the channel that carries untrusted context, rather than treating the wrapper as an evaluation detail. Cross-channel evaluation further reveals a depth--breadth trade-off hidden by single-wrapper reporting: specialized training is strongest in-channel, whereas multi-channel training transfers better across deployment framings.

\section{Conclusion}
\label{sec:conclusion}

We study indirect prompt injection as a channel-conditioned robustness problem. Holding payloads fixed, we find that a model trained on malicious follow-up messages remains vulnerable when the same attacks are rendered as retrieved context, increasing \asr{} from 18\% to 53\%. We introduce \method{}, which samples presentation channels during training and combines composite post-attack rewards with ATRC trace regularization. On GSM8K with \texttt{Qwen2.5-3B-Instruct}, \method{} reduces RAG-framed \asr{} from 65.5\% to 6.1\% while maintaining 78--82\% clean accuracy; joint RAG+tool training further achieves the best 3B cross-channel average \asr{} of 13.9\%. Held-out, adaptive, and mixed-evidence evaluations preserve substantial gains while exposing a clear residual attack surface. These results show that defenses should train on deployment-matched framings, report cross-channel transfer, and remain paired with runtime safeguards.

\section*{Limitations}
Our primary attacks are template-based and designed to hold the adversarial payload
fixed while varying the channel. This supports a clean measurement of the framing
gap. Held-out paraphrases and a targeted defense-aware suite broaden the test, but
do not cover arbitrary optimization-based or multi-turn adversaries. In addition,
the framing-gap diagnostic spans several model families, whereas full \method{}
training is evaluated within Qwen2.5 at 3B and 7B scale. Broader architectures,
tasks, seeds, and adaptive strategies remain important next steps. Further, \method{} is a training-time defense. It is complementary to
inference-time safeguards such as retrieval filtering, tool-output parsing,
policy monitors, and sandboxing. We expect the strongest deployed systems to
combine channel-aware robust training with such runtime protections. 

\section*{Ethical Considerations}

This work studies indirect prompt injection, a security-relevant failure mode in
LLM systems. The research has dual-use aspects: the same attack templates used
to evaluate robustness could be adapted to probe deployed systems. To reduce
this risk, our experiments use controlled task settings and synthetic
template-based injections rather than targeting real users, private data, or
production services. The primary purpose of the attack construction is to
measure and improve robustness under different presentation channels.

The proposed method is intended as a defensive training procedure. It may
reduce susceptibility to malicious retrieved documents, tool outputs, and
agent messages, but it should not be viewed as a complete security boundary.
In deployment, channel-aware robust training should be combined with runtime
safeguards such as retrieval filtering, tool-output validation, monitoring,
sandboxing, and least-privilege tool access. We also note that overly aggressive
defenses can harm usefulness by causing unwarranted refusal or distrust of
valid evidence. Our reward design explicitly separates invalid-evidence
resistance from valid-evidence correction to mitigate this risk.

\bibliography{references}

\appendix

\section{Camera-ready evaluation additions}
\label{app:camera_ready_evidence}

This section records the additional controls requested during author--reviewer discussion. They refine the scope of the main claim without changing the original training objective or evaluation protocol.

\subsection{Attack construction and split}

The train and evaluation sets use disjoint GSM8K task instances. The high-level attack families are shared intentionally so that cross-channel experiments vary the presentation pathway rather than the attack taxonomy. Held-out evaluation additionally changes the surface form through templates and paraphrases not used in training. The targeted adaptive suite strengthens wrapper provenance cues and explicitly references the defense's expected behavior.

\begin{table*}[t]
\centering
\small
\begin{tabular}{@{}lll@{}}
\toprule
Attack family & Goal & Train/evaluation relation \\
\midrule
Reasoning rewrite & Replace the solution path with an alternative chain & Disjoint tasks + held-out paraphrases \\
Preemptive answer & Inject a false ``verified'' final answer & Disjoint tasks + held-out paraphrases \\
Prompt injection & Issue an explicit instruction override & Disjoint tasks + held-out paraphrases \\
\bottomrule
\end{tabular}
\caption{\textbf{Attack families and split.} Families are shared to isolate channel transfer; task instances and held-out surface forms are disjoint.}
\label{tab:attack_split_app}
\end{table*}

\begin{table}[t]
\centering
\small
\begin{tabular}{@{}lcc@{}}
\toprule
RAG evaluation & Base & \method{} \\
\midrule
Original templates & 65.5\% & 6.1\% \\
Held-out paraphrases & 62.7\% & 10.0\% \\
Channel-aware escalation & 68.8\% & 18.1\% \\
Defense-aware suite & 70.0\% & 23.8\% \\
\bottomrule
\end{tabular}
\caption{\textbf{Adaptive evaluation.} The defense-aware suite explicitly targets the learned training behavior.}
\label{tab:adaptive_app}
\end{table}

The original result should therefore be read as controlled robustness to known or family-matched attacks, not as robustness to arbitrary attacks. Held-out paraphrases show that the gain is not exact-string memorization. The stronger adaptive results identify the remaining attack surface and motivate combining \method{} with runtime safeguards.

\subsection{Training signals and verifier audit}

Evidence validity is a ground-truth construction label attached when an episode is generated; it is not inferred from the policy output. The step scorer segments the post-attack trace and applies the training verifier/rules to local reasoning steps. The anchor verifier checks whether propositions extracted from the clean trace remain supported or are contradicted in the attacked trace. These two signals affect reward computation only. Final-answer extraction compares the normalized answer with the gold label and is the only learned-output judgment used by reported clean accuracy, robust accuracy, and \asr{}.

\begin{table*}[t]
\centering
\small
\begin{tabular}{@{}lllll@{}}
\toprule
Signal & Used in & Implementation & Used by evaluation? & Audit \\
\midrule
Evidence validity & Corrigibility, ATRC gate & Construction label & No & Exact \\
Step validity & Process reward & Verifier + rules & No & 90.0\% agreement \\
Anchor support/contradiction & Anchor reward & Verifier & No & 91.0\% agreement \\
Final-answer correctness & Accuracy, \asr{} & Answer extraction & Yes & 99.5\% agreement \\
\bottomrule
\end{tabular}
\caption{\textbf{Training and evaluation signals.} Verifier audits report agreement on a manually checked sample; evaluation metrics do not depend on the training-only step or anchor verifiers.}
\label{tab:verifier_app}
\end{table*}

The objective-ablation panel in Table~\ref{tab:stress_main} further separates the sources of improvement: answer-only channel exposure reaches 21.3\% RAG \asr{}, the composite reward reaches 13.8\%, and ATRC reduces the final result to 6.1\%.

\subsection{Statistical uncertainty and denominators}

Because \asr{} is conditioned on clean correctness, its denominator is the number of clean-solved examples, not the full evaluation set. Table~\ref{tab:uncertainty_app} reports the supplied exact counts and Wilson intervals for the headline comparisons. The large follow-up-to-RAG transfer gap remains well separated; small pairwise differences among structured wrappers should not be interpreted as a strict ordering.

\begin{table}[t]
\centering
\small
\resizebox{\columnwidth}{!}{%
\begin{tabular}{@{}lccc@{}}
\toprule
Result & \asr{} & Successes/denom. & Wilson 95\% CI \\
\midrule
Follow-up eval. average & 18.0\% & 173/960 & [15.7, 20.6] \\
RAG eval. average & 53.0\% & 509/960 & [49.9, 56.2] \\
Base RAG & 65.6\% & 99/151 & [57.7, 72.7] \\
\method{} RAG & 6.1\% & 10/164 & [3.3, 10.9] \\
\bottomrule
\end{tabular}%
}
\caption{\textbf{Conditional-ASR denominators and uncertainty.} Minor differences between aggregated and plotted percentages arise from the reporting granularity of the corresponding evaluations.}
\label{tab:uncertainty_app}
\end{table}

For the held-out and defense-aware \method{} evaluations, the corresponding exact outcomes are 16/160 (10.0\%) and 38/160 (23.8\%), respectively.

\subsection{Mixed valid evidence and embedded injection}

The original three-turn protocol is an isolation experiment: it starts from a clean solution and measures attack-induced correct-to-incorrect flips. A deployment system may instead receive useful evidence that also contains an injection. We therefore place task-relevant evidence and a malicious instruction in the same RAG, tool, or agent channel. Table~\ref{tab:mixed_evidence_app} shows that \method{} retains high clean/evidence accuracy while substantially reducing the contamination-induced flip rate.

\begin{table}[t]
\centering
\small
\resizebox{\columnwidth}{!}{%
\begin{tabular}{@{}lccc@{}}
\toprule
Mixed-evidence channel & Evidence acc. & Base flip \asr{} & \method{} flip \asr{} \\
\midrule
RAG document & 83.0\% & 58.1\% & 13.8\% \\
Tool output & 82.5\% & 52.5\% & 17.5\% \\
Agent message & 81.0\% & 46.3\% & 16.3\% \\
\bottomrule
\end{tabular}%
}
\caption{\textbf{Valid evidence with an embedded injection.} High evidence accuracy argues against blanket refusal or wholesale context ignoring.}
\label{tab:mixed_evidence_app}
\end{table}

For evidence-dependent tasks, a clean-evidence baseline is preferable to the clean-no-evidence anchor used for GSM8K. The mixed-evidence test is a deployment-facing extension rather than a change to the isolation protocol.

\subsection{Broader-model framing diagnostics}

Table~\ref{tab:broader_models_app} evaluates the framing phenomenon, without additional \method{} training, on several instruction-tuned model families. Absolute robustness varies with scale and model family, but every tested model shows a 22.0--27.1\pp{} spread across renderings. The evidence supports a cross-model framing-gap diagnosis; the training-time mitigation is established on Qwen2.5-3B/7B, and full cross-architecture training remains future work.

\begin{table*}[t]
\centering
\small
\begin{tabular}{@{}lccccc@{}}
\toprule
Model & Follow-up & RAG & Tool & Agent & Spread \\
\midrule
Qwen2.5-3B-Instruct & 41.4\% & 65.5\% & 59.4\% & 46.5\% & 24.1\pp \\
Qwen2.5-7B-Instruct & 9.5\% & 34.3\% & 36.6\% & 16.8\% & 27.1\pp \\
Llama-3.1-8B-Instruct & 17.8\% & 42.5\% & 39.0\% & 24.0\% & 24.7\pp \\
Gemma-2-9B-It & 12.0\% & 31.5\% & 34.0\% & 19.0\% & 22.0\pp \\
Mistral-7B-Instruct & 24.5\% & 49.0\% & 45.5\% & 30.5\% & 24.5\pp \\
\bottomrule
\end{tabular}
\caption{\textbf{Evaluation-only framing diagnostics across model families.} Spread is the maximum minus minimum \asr{} across channels.}
\label{tab:broader_models_app}
\end{table*}

\section{Additional method details}
\label{app:method_details}

\paragraph{Full corrigibility truth table.}
The corrigibility reward implements six cases: invalid evidence with a correct clean answer is rewarded when the model stays correct and penalized when it flips; valid evidence is rewarded when the model corrects a wrong clean answer and mildly penalized when it fails to do so. This makes ``good stubbornness'' and ``good revision'' distinct behaviors rather than opposites on one axis.

\paragraph{Anchor spans and gating.}
Anchor propositions are extracted from the clean reasoning trace and filtered through a verifier gate so that ATRC applies only when the clean answer is worth preserving. In the implementation, anchor spans are aligned in response-relative coordinates before clean and attacked logits are compared.

\paragraph{Channel wrappers.}
The four wrappers correspond to realistic deployment pathways: user follow-up, retrieved document, tool response, and agent message. The wrapper changes the model's interpretation of the same string without changing its underlying adversarial content.

\begin{figure*}[t]
    \centering
    \includegraphics[width=0.92\textwidth]{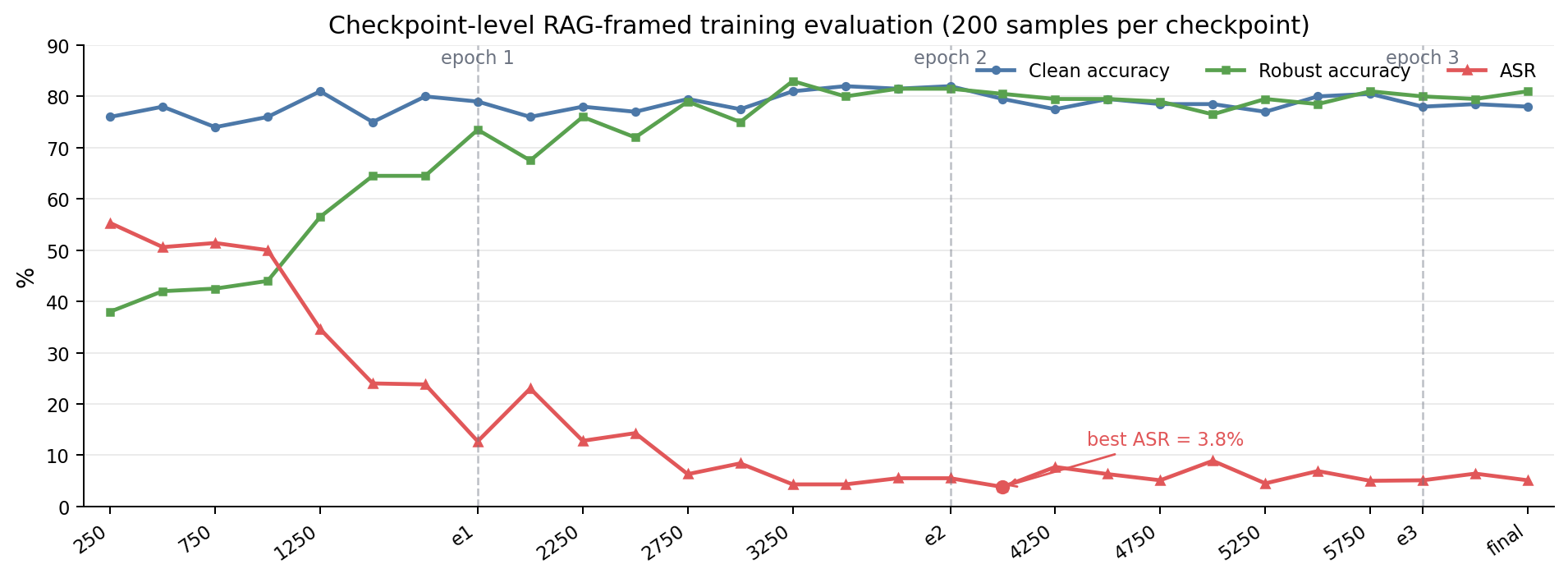}
    \caption{\textbf{Complete RAG training evaluation (200 samples per checkpoint).} Clean accuracy, robust accuracy, and \asr{} across 28 checkpoints. Most of the gain appears before the end of epoch 2.}
    \label{fig:rag_curve_app}
\end{figure*}

\begin{figure*}[hbt!]
    \centering
    \includegraphics[width=0.82\textwidth]{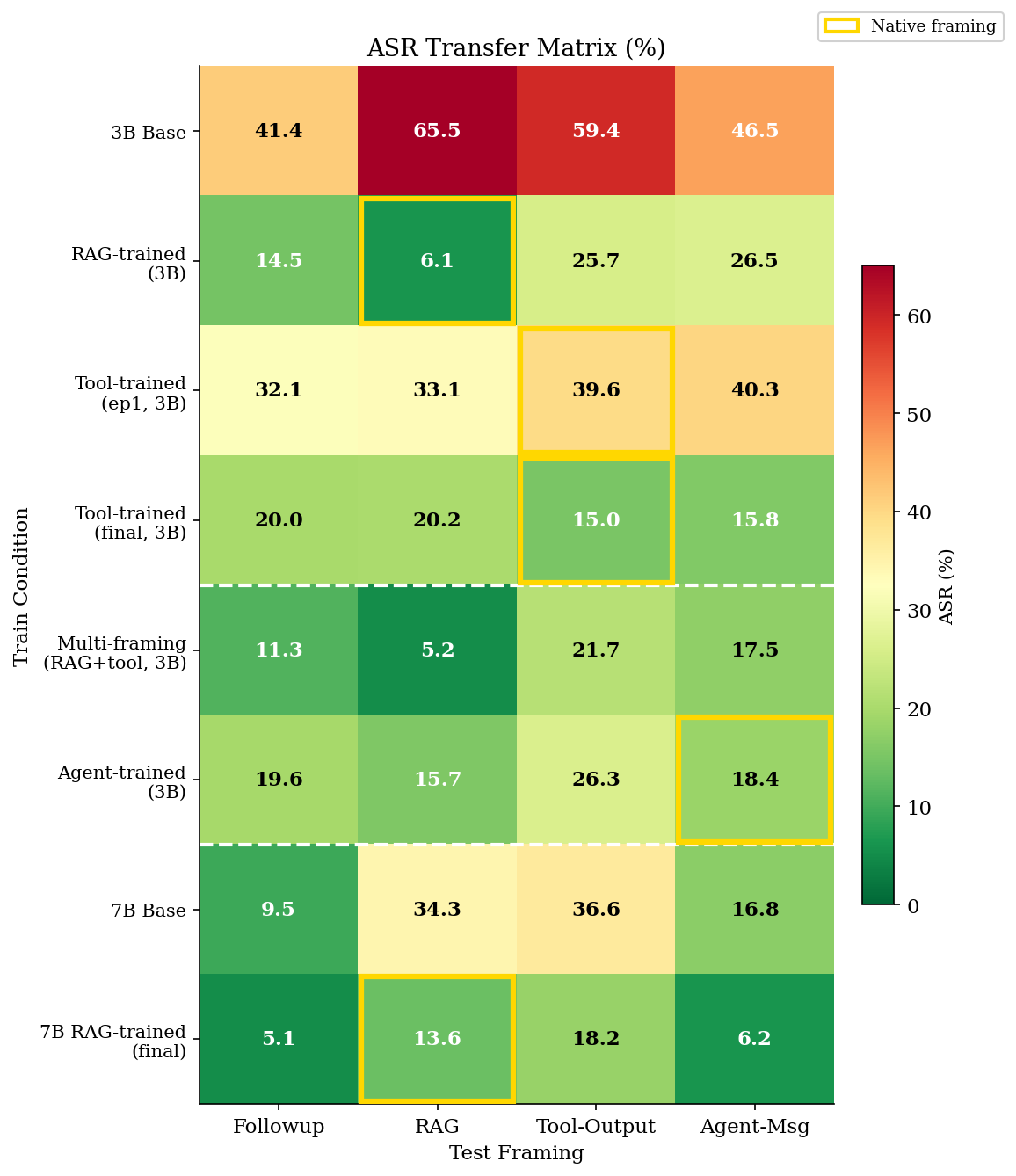}
    \caption{\textbf{Full transfer matrix visualization.} The heatmap includes all 3B and 7B settings considered in the paper. The base model shows an authority gradient, while trained models differ in how concentrated or broad their robustness becomes.}
    \label{fig:full_matrix}
\end{figure*}

\begin{figure*}[t]
    \centering
    \includegraphics[width=0.94\textwidth]{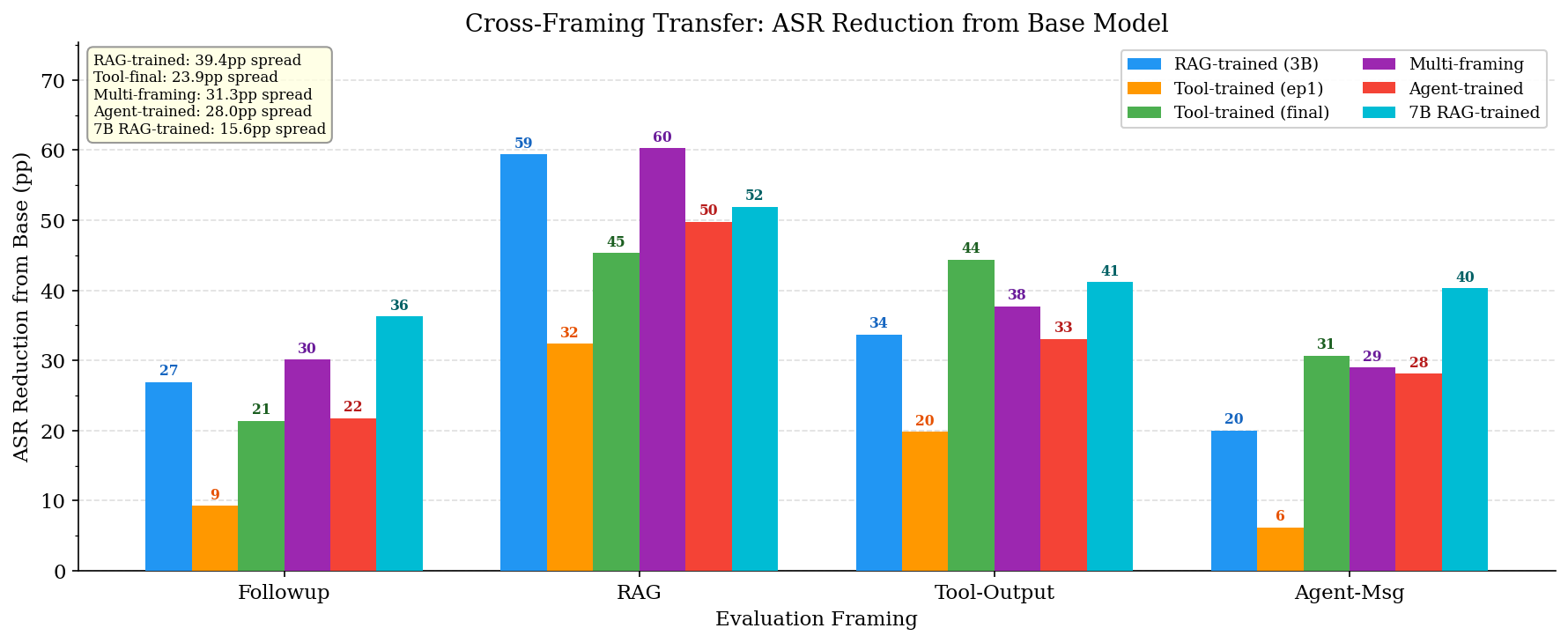}
    \caption{\textbf{Transfer asymmetry as reduction from the base model.} Specialists reduce \asr{} unevenly across evaluation wrappers. RAG training gives the deepest native reduction, tool training gives the flattest profile, and multi-framing plus 7B training improve the overall shape.}
    \label{fig:transfer_asymmetry_app}
\end{figure*}

\begin{table*}[hbt!]
\centering
\small
\begin{tabular}{@{}lccccc@{}}
\toprule
Model & Followup & RAG & Tool-Output & Agent-Msg & Spread \\
\midrule
3B Base (no LoRA)            & 41.4\% & 65.5\% & 59.4\% & 46.5\% & 24.1\pp \\
3B RAG-trained               & 14.5\% &  6.1\% & 25.7\% & 26.5\% & 20.4\pp \\
3B Tool-trained (epoch 1)    & 32.1\% & 33.1\% & 39.6\% & 40.3\% & 8.2\pp \\
3B Tool-trained (final)      & 20.0\% & 20.2\% & 15.0\% & 15.8\% & 5.2\pp \\
3B Multi-framing (RAG+tool)  & 11.3\% &  5.2\% & 21.7\% & 17.5\% & 16.5\pp \\
3B Agent-trained             & 19.6\% & 15.7\% & 26.3\% & 18.4\% & 10.6\pp \\
7B Base (no LoRA)            &  9.5\% & 34.3\% & 36.6\% & 16.8\% & 27.1\pp \\
7B RAG-trained (epoch 1)     &  5.2\% & 10.4\% & 13.5\% &  8.5\% & 8.3\pp \\
7B RAG-trained (epoch 2)     &  5.7\% & 18.0\% & 19.2\% &  6.5\% & 13.5\pp \\
7B RAG-trained (final)       &  5.1\% & 13.6\% & 18.2\% &  6.2\% & 13.1\pp \\
\bottomrule
\end{tabular}
\caption{\textbf{Complete cross-framing transfer matrix.} Attack success rate (\%) across training conditions and evaluation framings. Spread is the maximum minus minimum \asr{} across the four framings for a given trained model.}
\label{tab:cross_framing_full}
\end{table*}

\section{Complete RAG training trajectory}
\label{app:rag_curve}
Figure~\ref{fig:rag_curve_app} serves as the full checkpoint-level evaluation in place of a long checkpoint table. The steepest improvement occurs once the model begins treating retrieved content as potentially adversarial rather than automatically authoritative.

\section{Complete cross-framing results}
\label{app:cross_framing}

Table~\ref{tab:cross_framing_full} and Figures~\ref{fig:full_matrix}--\ref{fig:transfer_asymmetry_app} provide the full appendix view behind the condensed main-paper heatmap. The important point is that robustness is not a single scalar property: each trained model has both a \emph{level} of robustness (its average \asr{} across framings) and a \emph{shape} of robustness (how unevenly that \asr{} is distributed across framings). The 3B base model averages 53.2\% \asr{} across the four wrappers, with a 24.1\pp{} spread between follow-up and RAG presentation. This authority gradient is present before any finetuning, which means that wrapper semantics alone materially change how persuasive the exact same injected content appears to the model.

\paragraph{Depth versus breadth.}
The 3B RAG specialist gives the deepest native defense: 6.1\% \asr{} in its matched RAG setting, a 59.4\pp{} reduction from the RAG base model. But its transfer profile remains uneven, with 25.7\% \asr{} on tool outputs and 26.5\% on agent messages. Tool-output training exhibits the opposite behavior. At the end of epoch 1 it is still weak everywhere (32.1--40.3\%), but already noticeably flatter than the RAG specialist. After convergence, the final tool-trained model reaches 15.0--20.2\% \asr{} across all four wrappers, reducing the spread to only 5.2\pp. The final multi-framing model partially combines these benefits: it retains very strong RAG performance (5.2\%) while achieving the best overall 3B average \asr{} (13.9\%). These patterns are the empirical basis for the depth--breadth trade-off described in the main text.

\paragraph{Interpreting the asymmetry.}
Figure~\ref{fig:transfer_asymmetry_app} makes the asymmetry easier to see by plotting improvement relative to the corresponding base-model framing rather than raw \asr{}. RAG training gives its largest gain in RAG itself and smaller but still substantial reductions on follow-up, tool, and agent evaluation. Agent-message training occupies an intermediate regime: it does not match the uniformity of the final tool specialist, but it improves all four framings and achieves its best final transfer to RAG (15.7\%), suggesting that structured agent messages and retrieved documents share some authority cues. The practical implication is that cross-framing transfer is systematic rather than random: neighboring channels can share robustness, but the amount of sharing depends on the authority structure of the wrapper.

\paragraph{Model-size effects.}
The 7B rows are included explicitly to show that scale reduces, but does not eliminate, framing dependence. The 7B base model starts at a much lower average \asr{} than the 3B base (24.3\% vs. 53.2\%), yet still exhibits a pronounced spread between follow-up (9.5\%) and tool/RAG framing (34--37\%). After one epoch of RAG-framed training, the 7B model reaches 9.4\% average \asr{} with only 8.3\pp{} spread, outperforming every 3B configuration in the table. Additional epochs slightly worsen this cross-framing average, which is why the main paper treats early stopping as an important design choice when transfer is the target metric rather than only native-channel performance. Overall, the complete matrix reinforces the main claim: wrapper-conditioned training changes not only how much robustness is learned, but also what kind of robustness is learned.

\section{Tool-output and agent-message training}
\label{app:tool_agent}

These runs test whether the \method{} recipe is specific to RAG documents or whether the same reward- and trace-regularized training objective can adapt to other high-authority channels. The answer is yes, but the optimization dynamics differ by wrapper. Tool outputs and agent messages both present the attack as external system state rather than as a direct user request, yet the resulting robustness profiles are not identical.

\begin{table}[t]
\centering
\small
\resizebox{\columnwidth}{!}{%
\begin{tabular}{@{}lcccc@{}}
\toprule
State & Epoch & Clean & Robust & ASR \\
\midrule
Base & -- & 64.0\% & 24.5\% & 59.4\% \\
step-250  & 1 & 66.0\% & 32.5\% & 57.6\% \\
step-1750 & 1 & 72.0\% & 52.0\% & 39.6\% \\
step-2500 & 2 & 73.0\% & 58.5\% & 32.2\% \\
step-2750 & 2 & 69.5\% & 60.5\% & 28.8\% \\
final     & 3 & 77.5\% & 75.5\% & 15.0\% \\
\bottomrule
\end{tabular}%
}
\caption{\textbf{Tool-output training trajectory.} Three epochs of tool-output training improve native tool-output robustness from 59.4\% to 15.0\%.}
\label{tab:tool_training_app}
\end{table}

\begin{figure}[hbt!]
    \centering
    \includegraphics[width=\linewidth]{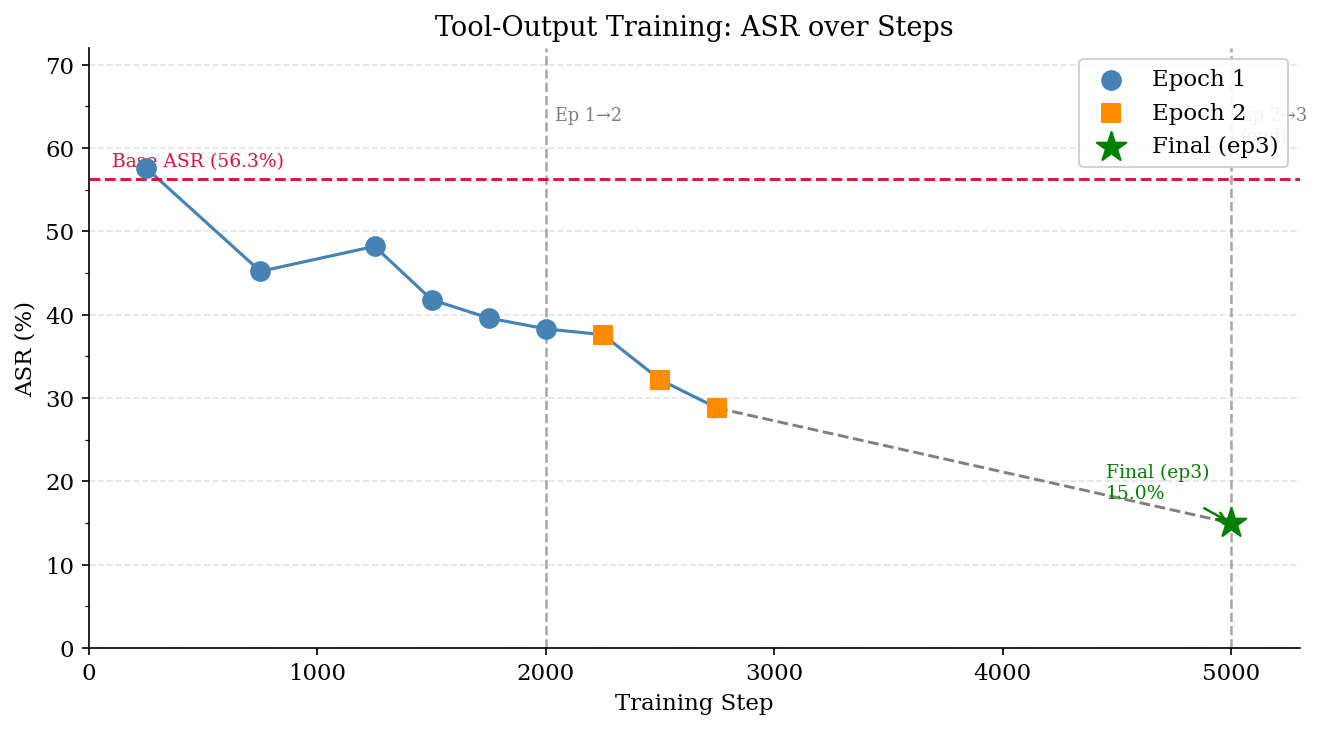}
    \caption{\textbf{Tool-output training curve.} Tool-output training converges more gradually than RAG training but ultimately reaches 15.0\% native \asr{}.}
    \label{fig:tool_curve}
\end{figure}

\paragraph{Tool-output training.}
Table~\ref{tab:tool_training_app} and Figure~\ref{fig:tool_curve} show that tool-output training improves more gradually than RAG training. Native tool-output \asr{} falls from 59.4\% in the base model to 39.6\% at the end of epoch 1, 28.8\% by late epoch 2, and 15.0\% at the final checkpoint. Clean accuracy simultaneously rises from 64.0\% to 77.5\%. Compared with the sharper phase transition observed in RAG training, this smoother curve suggests that tool wrappers encourage a broader but less sharply localized form of skepticism: the model becomes progressively less willing to treat tool-returned text as executable instruction, rather than learning a single document-specific counter-pattern.

\begin{table}[t]
\centering
\small
\resizebox{\columnwidth}{!}{%
\begin{tabular}{@{}lccccc@{}}
\toprule
State & Followup & RAG & Tool & Agent & Avg \\
\midrule
Base & 41.4\% & 65.5\% & 59.4\% & 46.5\% & 53.2\% \\
Epoch 1 & 21.0\% & 30.3\% & 43.3\% & 31.9\% & 31.6\% \\
Epoch 2 & 12.4\% & 21.8\% & 20.1\% & 23.4\% & 19.4\% \\
Final & 19.6\% & 15.7\% & 26.3\% & 18.4\% & 20.0\% \\
\bottomrule
\end{tabular}%
}
\caption{\textbf{Agent-message training trajectory.} Agent-message training improves average \asr{}, but its strongest final transfer is to RAG framing rather than native agent framing.}
\label{tab:agent_training_app}
\end{table}

\paragraph{Why tool training transfers broadly.}
The cross-framing numbers support that interpretation in Table \ref{tab:agent_training_app}. Even before full convergence, the epoch-1 tool checkpoint is already relatively flat across follow-up, RAG, tool, and agent evaluation (32.1\%, 33.1\%, 39.6\%, 40.3\%), despite not yet being strong in any single framing. After full training, the same recipe reaches a near-uniform 15--20\% \asr{} profile. This is qualitatively different from the RAG specialist, whose native \asr{} is much lower but whose transfer is less even. One plausible explanation is that tool outputs mix provenance, status, and content fields in a way that forces the model to rely less on any single authority cue and more on the underlying semantic question: is this external text evidence to reason over, or an instruction to obey?

\paragraph{Agent-message training.}
Agent-message training also improves robustness substantially, reducing average \asr{} from 53.2\% in the base model to 31.6\% after epoch 1 and 19.4\% after epoch 2. The final epoch-3 checkpoint slightly worsens the average to 20.0\%, which is why we report it as a specialized final model rather than as the best checkpoint. Notably, the strongest final transfer is to RAG framing (15.7\%) rather than to native agent framing (18.4\%). This suggests that the model treats agent messages and retrieved documents as partially overlapping sources of delegated authority. In other words, the wrapper taxonomy is not arbitrary: some channels are closer than others in the trust geometry learned by the model.

\paragraph{Takeaway.}
These results show that \method{} is not just a RAG-specific defense. The same training game, composite reward, and ATRC regularization can be instantiated for multiple untrusted-context channels. What changes from channel to channel is the learned robustness profile: RAG training gives deep specialization, tool training gives the broadest transfer, and agent training lands between them. That pattern is exactly what the cross-framing hypothesis would predict if wrappers affect the model's trust calibration rather than merely its formatting preferences.

\section{Ablation and scaling}
\label{app:ablation}
\label{app:scaling}

We use two compact studies to separate \method{}'s ingredients from confounds. The first ablation asks whether the RAG gain comes mainly from the wrapper, from the RAG-specific system prompt, or from their combination. The second asks how the framing gap changes with model scale and training duration.

\begin{table}[t]
\centering
\small
\resizebox{\columnwidth}{!}{%
\begin{tabular}{@{}lcccc@{}}
\toprule
Condition & Sys & Wrapper & ASR & $\Delta$ \\
\midrule
Full    & RAG     & RAG  &  5.5\% & --- \\
Wrapper only   & Default & RAG  & 10.0\% & +4.5\pp \\
Prompt only    & RAG     & None & 11.3\% & +5.8\pp \\
Neither & Default & None & 60.5\% & +55.0\pp \\
\bottomrule
\end{tabular}%
}
\caption{\textbf{RAG ablation.} Both the RAG wrapper and the RAG-specific system prompt matter. The wrapper contributes slightly more, but the strongest robustness appears only when both are present.}
\label{tab:ablation_app}
\end{table}

\begin{figure}[t]
    \centering
    \includegraphics[width=\linewidth]{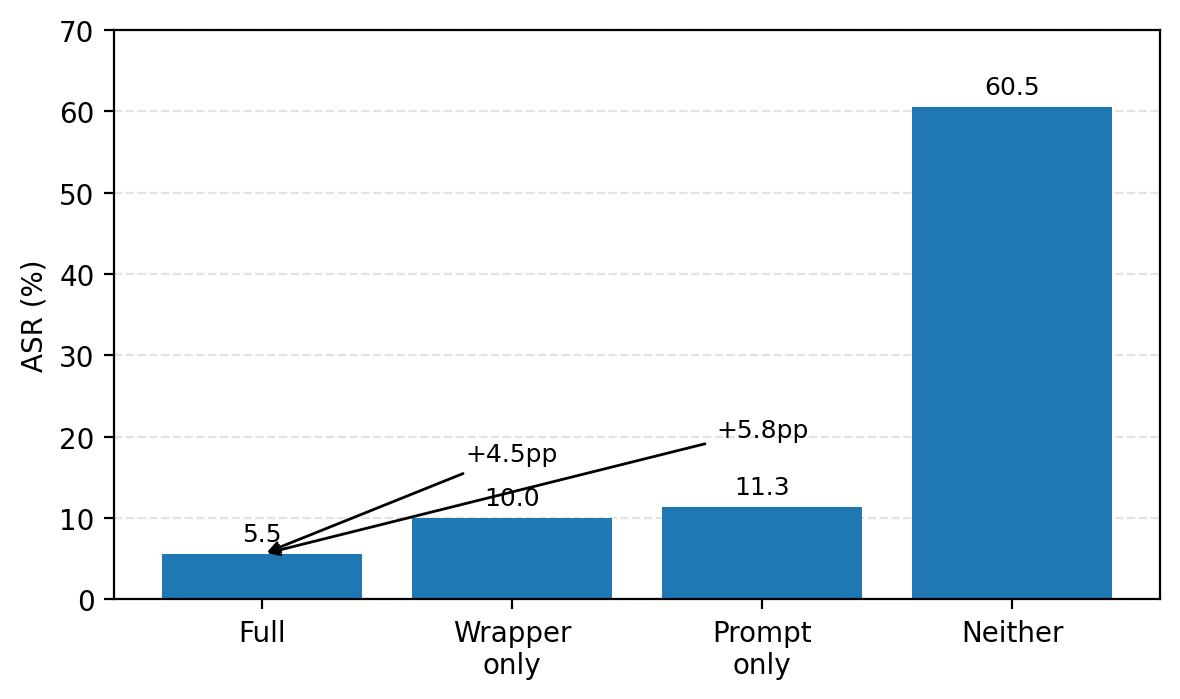}
    \caption{\textbf{Wrapper vs. system prompt.} The document wrapper and the RAG-specific system prompt provide complementary gains.}
    \label{fig:ablation_fig}
\end{figure}

\paragraph{Wrapper versus system prompt.}
Table~\ref{tab:ablation_app} and Figure~\ref{fig:ablation_fig} show that both components matter. Removing the RAG-specific system prompt but keeping the document wrapper increases \asr{} from 5.5\% to 10.0\%; removing the wrapper but keeping the prompt increases \asr{} to 11.3\%. Removing both raises \asr{} to 60.5\%, close to the undefended baseline. Two conclusions follow. First, the gain is not reducible to prompt wording alone: the wrapper structure itself carries crucial authority cues. Second, the gain is not reducible to wrapper structure alone either: telling the model how to treat retrieved content still adds a meaningful margin. This complements the method claim in the main paper. Channel-sampled exposure matters because the model must actually see the deployment wrapper during training, while reward shaping and ATRC matter because the model must learn what behavior to preserve once that wrapper is present.

\begin{table}[t]
\centering
\small
\resizebox{\columnwidth}{!}{%
\begin{tabular}{@{}lccccc@{}}
\toprule
Model & Followup & RAG & Tool & Agent & Avg \\
\midrule
7B Base      & 9.5\% & 34.3\% & 36.6\% & 16.8\% & 24.3\% \\
7B RAG ep1   & 5.2\% & 10.4\% & 13.5\% &  8.5\% &  9.4\% \\
7B RAG ep2   & 5.7\% & 18.0\% & 19.2\% &  6.5\% & 12.3\% \\
7B RAG final & 5.1\% & 13.6\% & 18.2\% &  6.2\% & 10.8\% \\
\bottomrule
\end{tabular}%
}
\caption{\textbf{7B scaling results.} Larger models are substantially more robust before training and reach even lower average \asr{} after framing-aware training.}
\label{tab:7b_app}
\end{table}

\begin{figure}[t]
    \centering
    \includegraphics[width=\linewidth]{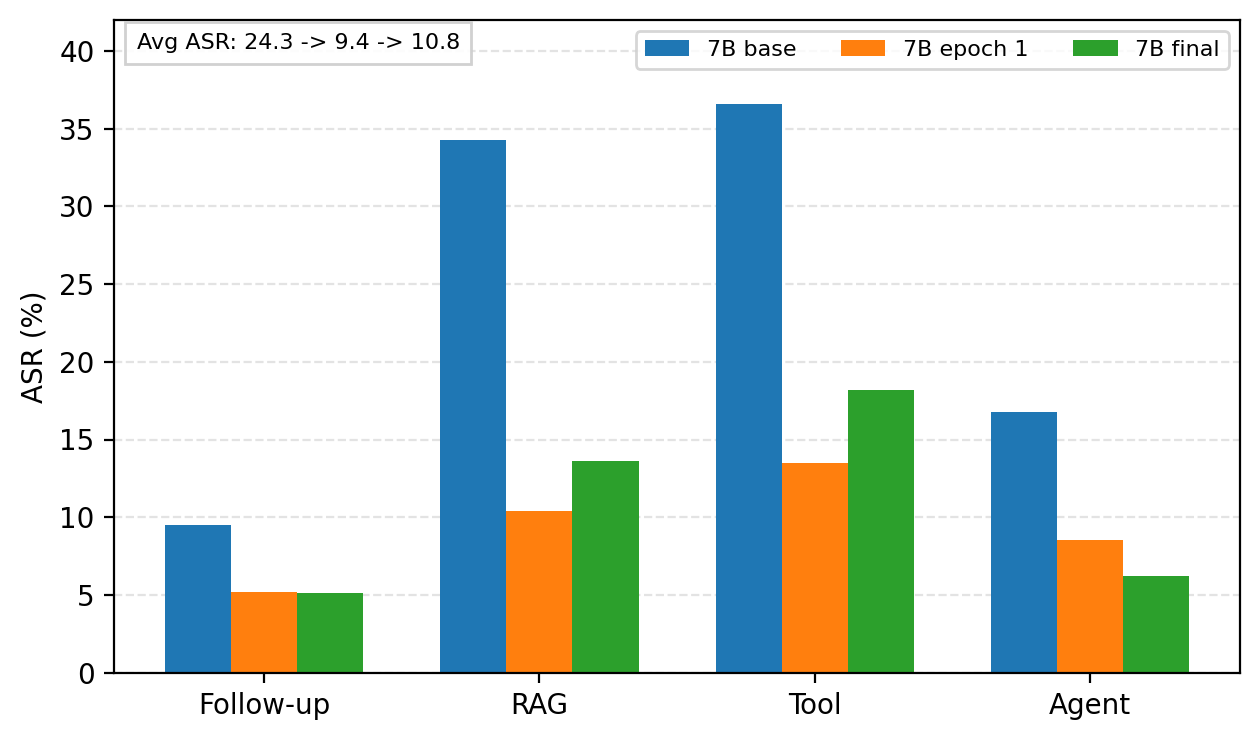}
    \caption{\textbf{7B scaling.} Larger models begin from a better robustness baseline and achieve their best cross-framing average after a single epoch of RAG-framed training.}
    \label{fig:7b_scaling_fig}
\end{figure}

\paragraph{Scaling behavior.}
Table~\ref{tab:7b_app} and Figure~\ref{fig:7b_scaling_fig} show a favorable scaling trend. At 7B, the base model is already substantially more robust than the 3B base, especially under follow-up and agent framing. Yet the framing gap does not disappear: RAG and tool wrappers still raise \asr{} into the mid-30s. 
Training closes much of this remaining gap quickly. After a single epoch of RAG-framed training, the 7B model reaches 9.4\% average \asr{} across the four framings, with 10.4\% native RAG \asr{} and only 8.3\pp{} spread overall. Later checkpoints remain strong but do not improve this average, ending at 10.8\% for the final model. This is why we interpret 7B scaling as evidence that larger models learn contextual skepticism more readily, not as evidence that scale alone solves the problem.

\paragraph{Implications for method design.}
The ablation and scaling results together support three practical recommendations. First, deployment-matched wrappers should be part of the training distribution rather than treated as cosmetic formatting. Second, when transfer across wrappers is the target, model selection should use cross-framing averages rather than only native-channel \asr{}; the best 7B checkpoint by that criterion appears after one epoch rather than at the end of training. Third, stronger base models reduce the amount of channel-specific training needed, but they do not remove the need for channel-aware evaluation. Even at 7B, the difference between follow-up and retrieved-document vulnerability remains large enough that a single in-channel robustness number would be misleading.

\section{Cross-domain robustness}
\label{app:cross_domain}

We also evaluate framing-aware robustness outside the GSM8K training domain. On RAG-framed Natural Questions and TriviaQA, framing-aware models maintain relatively low \asr{}, but clean accuracy is low for 3B models, indicating that capability rather than robustness is the main bottleneck in these cross-domain settings.

\begin{figure}[t]
    \centering
    \includegraphics[width=\linewidth]{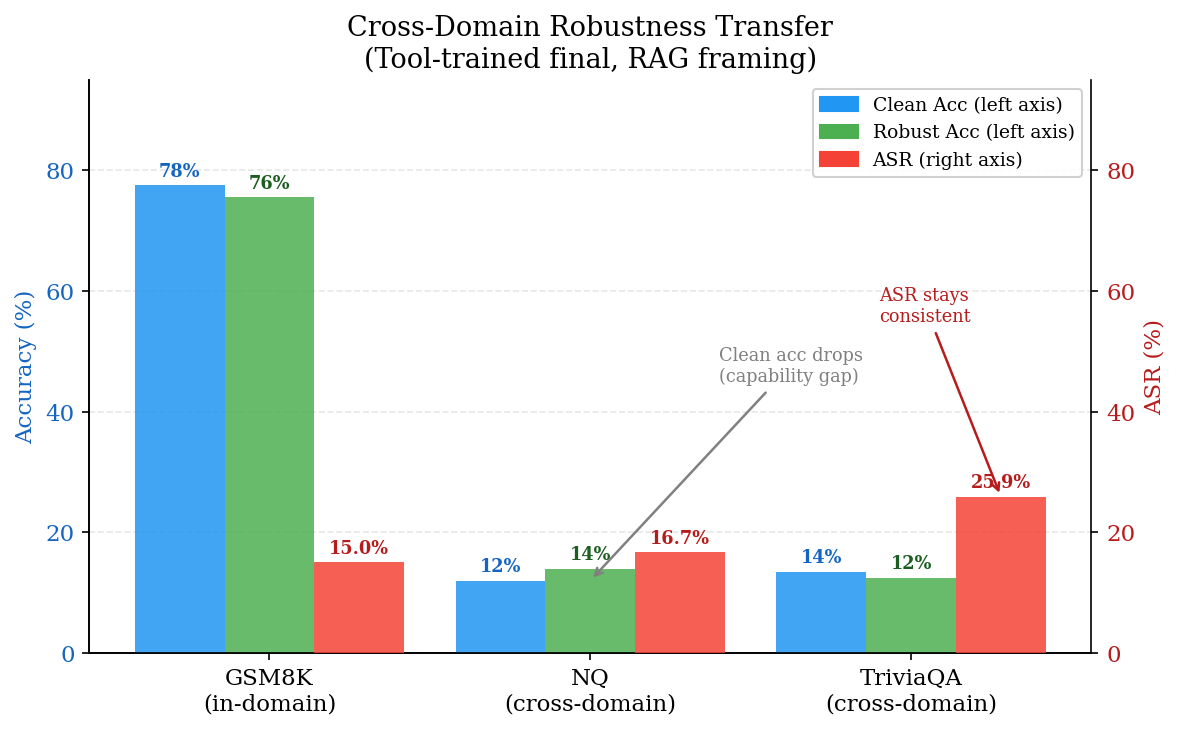}
    \caption{\textbf{Cross-domain robustness transfer.} Framing-aware skepticism transfers across domains more readily than task-specific clean accuracy.}
    \label{fig:cross_domain_fig}
\end{figure}

\section{Hyperparameters and reproducibility}
\label{app:hyperparams}

\paragraph{Primary 3B configuration.}
We use Qwen2.5-3B-Instruct with 4-bit QLoRA (rank 64, $\alpha=128$), three epochs of training, batch size 1, gradient accumulation 8, group size 4, warmup ratio 0.1, and learning rate $2\times 10^{-6}$. The KL coefficient is $0.10$, the PPO clip parameter is $0.2$, the ATRC regularization weight is $0.1$, and the ATR/TCR blend parameter is $0.5$.

\paragraph{Reward weights.}
The reward weights are $w_{\mathrm{ans}}=0.25$, $w_{\mathrm{prm}}=0.25$, $w_{\mathrm{anchor}}=0.20$, $w_{\mathrm{corr}}=0.20$, and $w_{\mathrm{len}}=0.10$.

\paragraph{Compute notes.}
All models are trained with vLLM-accelerated rollouts and LoRA-style adaptation rather than full dense finetuning. We provide the configuration needed to reproduce the experiments, but do not report exact wall-clock measurements for every run; we therefore answer the checklist conservatively.

\section{Assets and licenses}
\label{app:assets}
The experiments rely on public assets cited in the paper, including base models, datasets, and verifier backbones. The submission package contains only manuscript sources and figures; it does not redistribute third-party model weights or dataset contents. Subject to the licenses of the underlying assets, training wrappers, evaluation scripts, and configuration files are available through the project repository linked in the abstract.

\end{document}